%% file: main.tex
\documentclass[10pt,twocolumn,letterpaper]{article}
\usepackage[letterpaper, top=1in, bottom=1in, left=1in, right=1in]{geometry}
\usepackage{cvpr}              
\usepackage{pifont}  
\usepackage{array}   
\usepackage{multirow}

\definecolor{cvprblue}{rgb}{0.21,0.49,0.74}
\usepackage[pagebackref,breaklinks,colorlinks,citecolor=cvprblue]{hyperref}

\def\paperID{9848} 
\def\confName{CVPR}
\def\confYear{2025}

\title{SDGOCC: Semantic and Depth-Guided Bird’s-Eye View Transformation for 3D Multimodal Occupancy Prediction}
\author{ZaiPeng Duan\hspace{0.2em} ChenXu Dang\hspace{0.2em}
Xuzhong Hu \hspace{0.2em} Pei An \hspace{0.2em} Junfeng Ding \hspace{0.2em}Jie Zhan \hspace{0.2em} YunBiao Xu \hspace{0.2em} \thanks{Corresponding author:  majie@hust.edu.cn} Jie Ma\\
Huazhong University of Science and Technology\\}

\begin{document}
\maketitle
\input{sec/0_abstract}    
\input{sec/1_intro}
\input{sec/2_related_work}

\input{sec/3_Methodology}
\input{sec/4_Experiments}

\input{sec/5_conclusion}


{
    \small
    \bibliographystyle{ieeenat_fullname}
    \bibliography{main}
}

\end{document}

%% file: sec/0_abstract.tex
\begin{abstract}
Multimodal 3D occupancy prediction has garnered significant attention for its potential in autonomous driving. However, most existing approaches are single-modality: camera-based methods lack depth information, while LiDAR-based methods struggle with occlusions. Current lightweight methods primarily rely on the Lift-Splat-Shoot (LSS) pipeline, which suffers from inaccurate depth estimation and fails to fully exploit the geometric and semantic information of 3D LiDAR points. Therefore, we propose a novel multimodal occupancy prediction network called SDG-OCC, which incorporates a joint semantic and depth-guided view transformation coupled with a fusion-to-occupancy-driven active distillation. The enhanced view transformation constructs accurate depth distributions by integrating pixel semantics and co-point depth through diffusion and bilinear discretization. The fusion-to-occupancy-driven active distillation extracts rich semantic information from multimodal data and selectively transfers knowledge to image features based on LiDAR-identified regions. Finally, for optimal performance, we introduce SDG-Fusion, which uses fusion alone, and SDG-KL, which integrates both fusion and distillation for faster inference. Our method achieves state-of-the-art (SOTA) performance with real-time processing on the Occ3D-nuScenes dataset and shows comparable performance on the more challenging SurroundOcc-nuScenes dataset, demonstrating its effectiveness and robustness. The code will be released at \url{https://github.com/DzpLab/SDGOCC}.
\end{abstract}

%% file: sec/1_intro.tex
\section{Introduction}
Accurate 3D perception of the surrounding environment forms the cornerstone of modern autonomous driving systems and robotics, ensuring efficient planning and safe control \cite{1,2}. In recent years, advancements in 3D object detection \cite{3,4,5,6,7} and semantic segmentation \cite{8,9,10,11,12} have significantly propelled the field of 3D perception. However, object detection relies on strict bounding boxes, making it difficult to recognize arbitrary shapes or unknown objects, while semantic segmentation struggles with fine-grained classification in complex scenes, especially under occlusion and overlap. In this context, 3D semantic occupancy prediction \cite{13,14} offers a more comprehensive approach to environment modeling. It simultaneously estimates the geometric structure and semantic categories of scene voxels, assigns labels to each 3D voxel, and provides a more complete perception, showing stronger robustness to arbitrary shapes and dynamic occlusions.

\begin{figure}[t]
  \centering
  \includegraphics[width=1.0\linewidth,height=!]{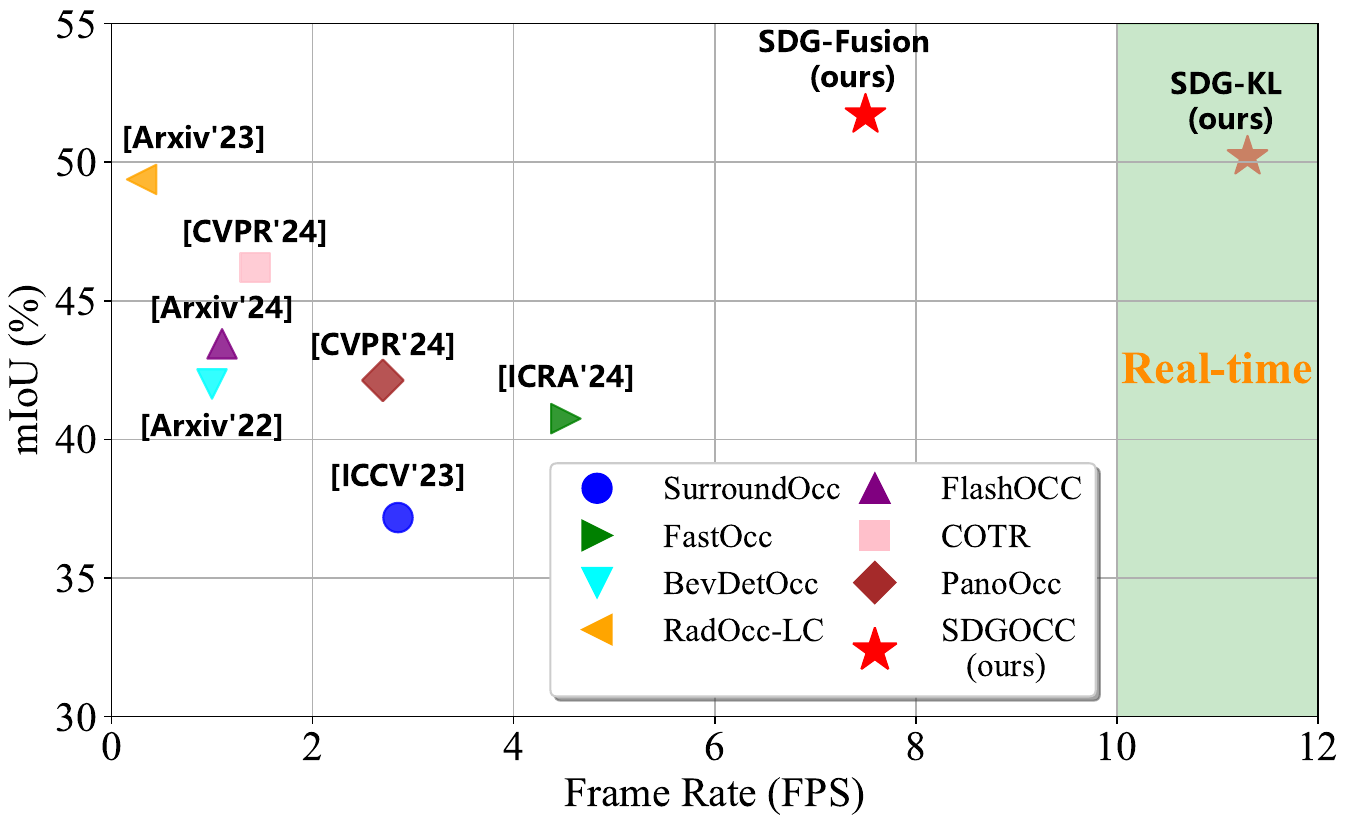}
  \vspace{-14pt}
  \caption{Comparisons of the mIoU and inference speed (FPS) of various 3D occupancy prediction methods on the Occ3D-nuScenes validation set. SDG-OCC achieves higher accuracy and real-time inference speed.}
  \label{fig:fps}
  \vspace{-12pt}
\end{figure}
\begin{figure}[t]
    \centering
    \begin{subfigure}{0.45\columnwidth} 
        \centering
        \includegraphics[width=\linewidth]{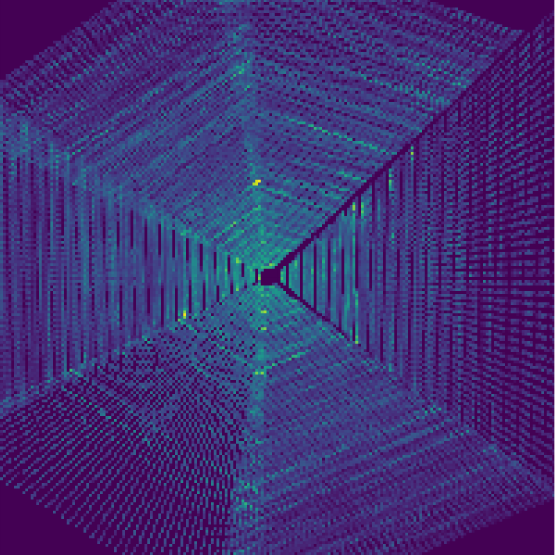}
        \caption{Initial BEV features of LSS}
    \end{subfigure}
    \hspace{2pt}
    \begin{subfigure}{0.45\columnwidth} 
        \centering
        \includegraphics[width=\linewidth]{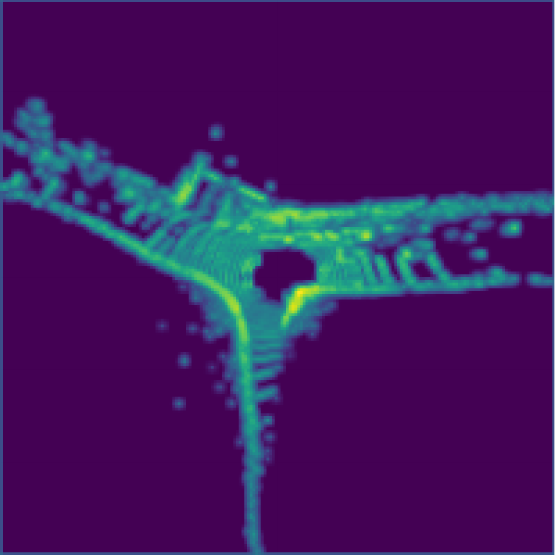}
        \caption{Our initial BEV features}
    \end{subfigure}
    \vspace{0.8em} 
    \begin{subfigure}{0.45\columnwidth} 
        \centering
        \includegraphics[width=\linewidth]{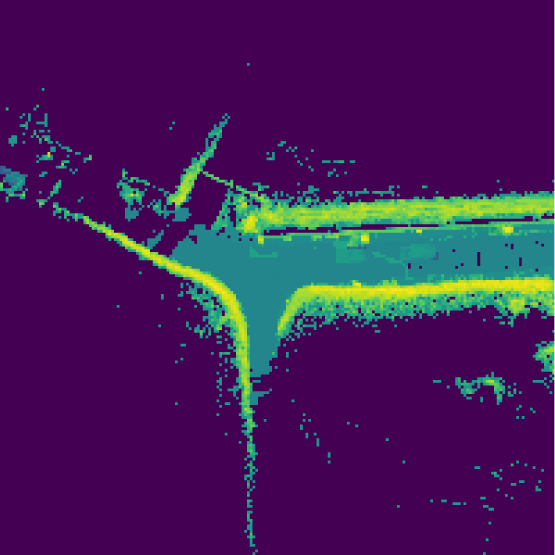}
        \caption{Ground truth of BEV}
    \end{subfigure}
    \hspace{1.8pt}
    \begin{subfigure}{0.45\columnwidth} 
        \centering
        \includegraphics[width=\linewidth]{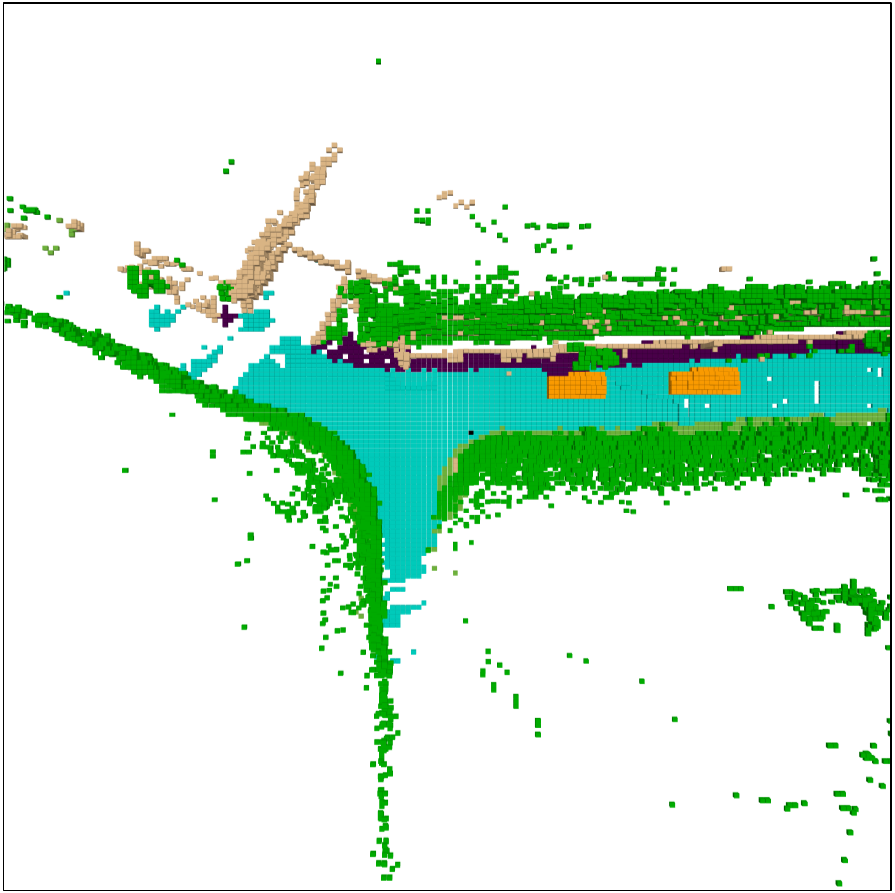}
        \caption{Occupancy grid of BEV}
    \end{subfigure}
\vspace{-10pt}
\caption{(a) BEV feature map of LSS with a shape of 200×200. We can observe that LSS has an extremely low utilization rate for BEV space. (b) The corresponding BEV features in SDG-OCC. Using depth and semantic information, the 2D-to-3D view transformation achieves efficient occupancy and utilization of the BEV features. (c) The corresponding BEV features in Ground Truth. (d) The corresponding BEV features in the occupancy grid.}
\vspace{-12pt} 
\label{fig:bev}
\end{figure}

Leveraging the complementary strengths of LiDAR and camera data is crucial for various 3D perception tasks. However, due to the heterogeneity between modalities, fusing LiDAR and camera data for 3D occupancy prediction remains challenging. Specifically, cameras provide rich semantic information but lack precise depth details, while LiDAR offers accurate depth information but only captures sparse data, potentially missing comprehensive scene details such as occluded objects. Existing methods often suffer from significant computational burdens (see \cref{fig:fps}), with some approaches attempting to leverage the LSS \cite{15} pipeline for real-time performance. Although LSS simulates the uncertainty of each pixel's depth through depth distribution (with depth intervals typically set to 0.5m), its sparse BEV representation allows only 50\% of the grids to receive valid image features \cite{16} (see \cref{fig:bev} (a)). While increasing the depth interval can improve depth estimation accuracy to mitigate sparsity, it significantly increases computational demands.Additionally, while LiDAR can provide valuable geometric priors, fusion-based methods that process both point clouds and images simultaneously impose heavy computational burdens, thereby increasing the strain on real-time applications.

To address these issues, we propose a multimodal 3D semantic occupancy prediction framework, named SDG-OCC, which aims to achieve higher accuracy and competitive inference speed by fusing LiDAR information in the BEV perspective. In this framework, we introduce a semantic and depth-guided view transformation to replace normal BEV feature generation. Specifically, after extracting features from camera data and obtaining semantic segmentation masks and depth distributions through a multi-task head, we use the semantic masks and depth maps provided by LiDAR to construct virtual points via local diffusion and bilinear discretization. Combined with the depth distribution, these points are then projected into the BEV space. The comparison between LSS and our generated BEV features is shown in \cref{fig:bev}. The SDG view transformation significantly refines depth estimation accuracy and reduces redundant virtual point seeds, improving both the speed and accuracy of semantic occupancy.

Secondly, we introduce a fusion-to-occupancy-driven active distillation module. We first fuse LiDAR and camera features in the BEV space and then unidirectionally selectively transferred multimodal knowledge to image features based on LiDAR-identified regions. Our proposed SDG-Fusion, which includes only fusion, achieved SOTA performance on the Occ3D-nuscenes \cite{13} and SurroundOcc-nuScenes \cite{24} validation dataset. In comparison, SDG-KL, which combines fusion and unidirectional distillation, achieves real-time speed with a slight performance penalty.

Our contributions can be summarized as follows:
\begin{itemize}
\item We introduce a multimodal 3D semantic occupancy prediction framework, termed SDG-OCC, aimed at achieving higher accuracy and competitive inference speed by fusing LiDAR information in the BEV perspective.

\item We propose a novel view transformation method that leverages the geometric and semantic information of point clouds to guide the 2D-3D view transformation. This significantly enhances the accuracy of depth estimation and improves both the speed and accuracy of semantic occupancy.

\item We propose a fusion-to-occupancy-driven active distillation module that integrates multimodal features and selectively transfers multimodal knowledge to image features based on LiDAR-identified regions. Building on this, we present SDG-Fusion for high performance and SDG-KL for faster inference.

\item Our method achieves SOTA performance with real-time processing on the Occ3D-nuScenes dataset and shows comparable performance on the more challenging SurroundOcc-nuScenes validation dataset, demonstrating the effectiveness of our approach.
\end{itemize}

%% file: sec/2_related_work.tex
\section{Related Work}
\label{sec:formatting}


\subsection{ Vision-Centric Occupancy Perception}
Inspired by Tesla's autonomous driving perception system, vision-centric occupancy perception has garnered significant attention from both industry and academia. MonoScene \cite{22} is a pioneering work that used only RGB inputs. TPVFormer \cite{23} combines surround multi-camera inputs and uses transformer-based methods to lift features into a tri-perspective view space. SurroundOcc \cite{24} extends high-dimensional BEV features into occupancy features and directly performs spatial cross-attention to generate geometric information. VoxFormer \cite{25} introduces a two-stage transformer-based semantic scene completion framework, capable of outputting complete 3D volumetric semantics from 2D images alone. FlashOcc \cite{26} transforms the channel to height, lifting BEV output to 3D space, significantly improving operational efficiency. FBOcc \cite{27} proposes a front-to-back view transformation module based on BEV features to address the limitations of different view transformations. Methods like UniOcc \cite{28} and RenderOcc \cite{29} use NeRF \cite{30} to directly predict 3D semantic occupancy, but the rendering speed limits their efficiency. FastOcc \cite{31} improves the occupancy prediction head to achieve a faster inference speed. COTR \cite{32} builds compact 3D occupancy representations through explicit-implicit view transformation and coarse-to-fine semantic grouping. In this paper, we improve the speed and accuracy of 3D semantic occupancy prediction from the BEV space by incorporating the geometric and semantic information of the point cloud into the view transformation.
\subsection{Multi-Modal Occupancy Perception}
Multimodal occupancy perception leverages the strengths of multiple modalities to overcome the limitations of unimodal perception. OpenOccupancy\cite{14} introduced a benchmark for LiDAR-camera semantic occupancy prediction. Inspired by BEVFusion, OccFusion \cite{33} concatenates 3D feature volumes from different modalities along the feature channels, followed by convolutional layers to combine them. CO-Occ \cite{34} introduced the Geometric and Semantic Fusion (GSFusion) module, identifying voxels containing both point cloud and visual information using k-nearest neighbors (KNN) search. OccGen \cite{35} employs an adaptive fusion module to dynamically integrate occupancy representations from camera and LiDAR branches, using 3D convolutions to determine fusion weights for aggregating LiDAR and camera features. HyDR \cite{36} proposes to integrate multimodal information in both perspective view (PV) and bird's-eye view (BEV) representation spaces. In this paper, we enhance view transformation by incorporating semantic segmentation masks and LiDAR depth maps to achieve higher occupancy accuracy. Additionally, we fuse BEV features from multimodal data and unidirectionally distill them into camera features, improving the accuracy and inference speed of 3D semantic occupancy prediction.

%% file: sec/3_Methodology.tex
\definecolor{nbarrier}{RGB}{112, 128, 144}
\definecolor{nbicycle}{RGB}{220, 20, 60}
\definecolor{nbus}{RGB}{255, 127, 80}
\definecolor{ncar}{RGB}{255, 158, 0}
\definecolor{nconstruct}{RGB}{233, 150, 70}
\definecolor{nmotor}{RGB}{255, 61, 99}
\definecolor{npedestrian}{RGB}{0, 0, 230}
\definecolor{ntraffic}{RGB}{47, 79, 79}
\definecolor{ntrailer}{RGB}{255, 140, 0}
\definecolor{ntruck}{RGB}{255, 99, 71}
\definecolor{ndriveable}{RGB}{0, 207, 191}
\definecolor{nother}{RGB}{175, 0, 75}
\definecolor{nsidewalk}{RGB}{75, 0, 75}
\definecolor{nterrain}{RGB}{112, 180, 60}
\definecolor{nmanmade}{RGB}{222, 184, 135}
\definecolor{nvegetation}{RGB}{0, 175, 0}
\definecolor{nothers}{RGB}{0, 0, 0}
\section{Methodology}
\subsection{Preliminary}
Given joint input from multi-view images and LiDAR data, 3D occupancy prediction aims to estimate the occupancy state and semantic classification of 3D voxels surrounding the ego vehicle. Specifically, the input consists of a $T$-frame consequent sequence of images ${X}_C \in \mathbb{R}^{N_C \times H_C \times W_C \times 3}$ from ${N_C}$ surround-view cameras and point clouds ${X}_L \in \mathbb{R}^{N_L \times (3 + d)}$ as multimodal input, represented as ${X}$=\{${X_C,X_L}$\}. Here ${H_C}$, ${W_C}$ represent the height and width of the image, respectively, ${N_L}$ denotes the number of point clouds and $d$ denotes the initial additional features of the point cloud. Subsequently, we train a neural network to generate an occupancy voxel map ${Y} \in \mathbb{R}^{H \times W \times D \times C_N}$, where each voxel is assigned a label as unknown, occupied, or a semantic category from \{$C_0 \text{ to } C_N$\}. Here, $N$ denotes the total number of categories of interest, and $H,W,D$ represent the volume dimensions of the entire scene. 
\begin{figure*}[t]
  \centering
  \includegraphics[width=0.99\linewidth,height=!]{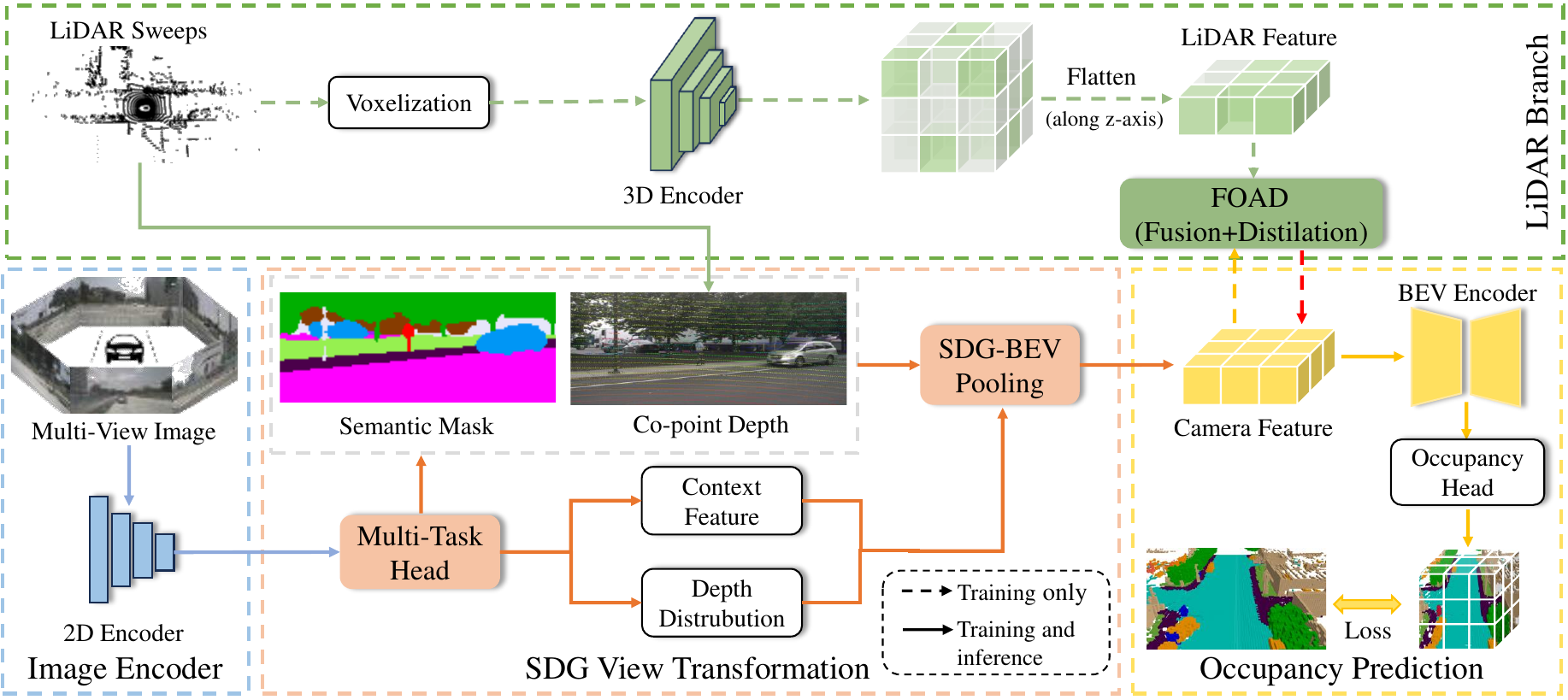}
  \vspace{-8pt}
  \caption{The overall architecture of SDG-OCC. $T$-frame multi-view images and corresponding point clouds are fed into the image and LiDAR backbones to extract features. Image features are processed by a multi-task head to generate semantic masks and depth distributions, combined with LiDAR depth maps to create virtual points for image BEV features. These features are fused with point cloud BEV features and selectively transfer multimodal knowledge to image features based on LiDAR-identified regions. Finally, enhanced features are processed by the occupancy prediction head to generate the occupancy map.}
  \label{fig:1}
  \vspace{-12pt}
\end{figure*}
\subsection{Overall Architecture}
An overview of SDGOCC is shown in \cref{fig:1}. It mainly consists of four key modules: image feature encoder to extract image features, semantic and depth-guided view transformation to construct 2D-3D feature transformation, fusion-to-occupancy-driven active distillation for fusing multimodal features and selectively transferring knowledge to the image features, and the occupancy prediction head for final output.  
\subsection{Image Encoder}
The image feature encoder aims to capture multi-view features, providing a foundation for 2D-3D view transformation. Given RGB images from surround-view cameras, we first use a pre-trained image backbone network, such as classic ResNet \cite{38} or strong Swin-Transformer \cite{40}, to extract multi-layer image features ${F}_C \in \mathbb{R}^{N_C \times C \times H \times W}$. These features are then aggregated using a feature pyramid network (FPN) \cite{39}, which combines fine-grained features and coarse-grained features and down-sampling them to a specific scale, typically 1/16.
\subsection{SDG View Transformation}
The LSS pipeline is widely used for converting image features to BEV representations in 3D perception. It constructs virtual points based on a predefined depth range for each pixel and predicts the depth distribution weight ${\alpha}$ and context feature $c$. The feature representation at depth $d$ is given by $p_d$ =${\alpha_d}c$. All virtual points then are projected into BEV space, where features at each height $Z$ are aggregated to form BEV features. Although LSS handles depth uncertainties and ambiguities by modeling pixel depth using depth distributions, the number of per-pixel features remains large even with a 0.5-meter depth interval, an order of magnitude larger than point features. Meanwhile, BEV features are highly sparse, with less than 50\% of the image features being effective, leading to suboptimal occupancy prediction performance. Reducing the depth interval improves accuracy but significantly increases computational burden and introduces irrelevant features, as most of the occupancy grid remains empty.
\begin{figure}[t]
  \centering
  \includegraphics[width=1.0\linewidth,height=!]{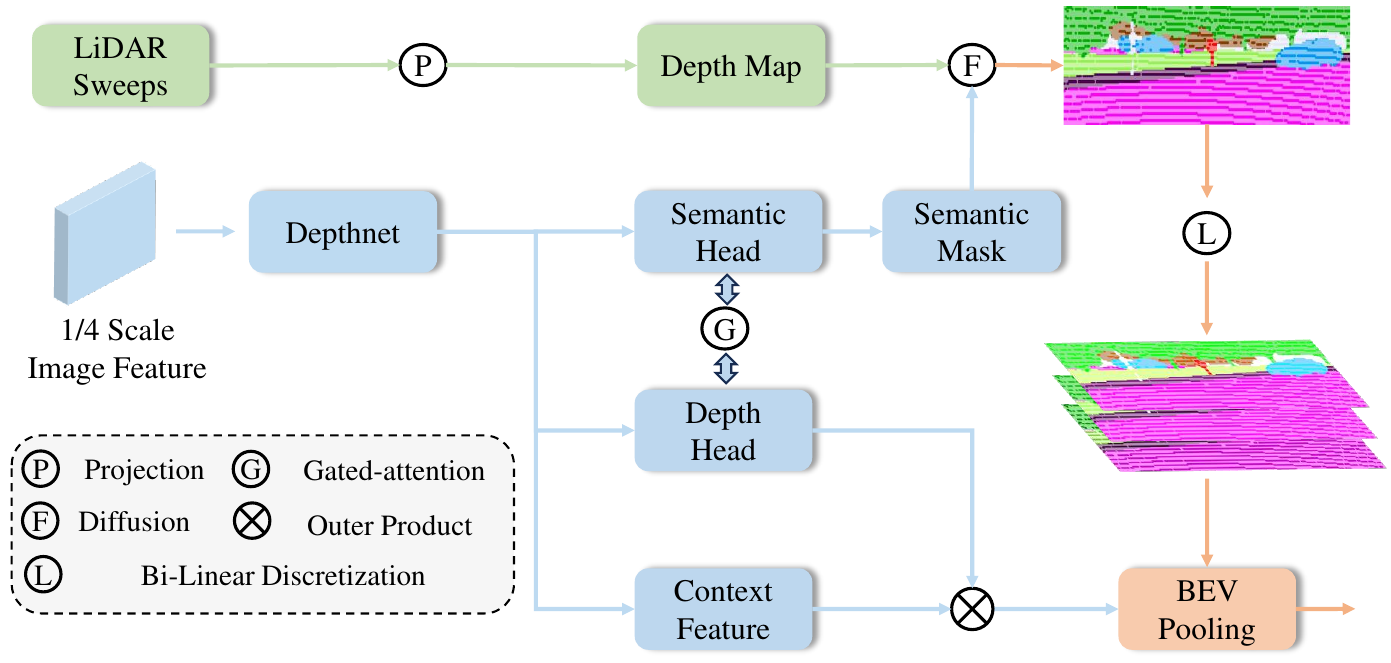}
  \vspace{-8pt}
  \caption{The overall architecture of the SDG view transformation. The image features processed by DepthNet are divided into texture features, depth features, and semantic features. Semantic features generate masks via a segmentation head, which are combined with depth maps from LiDAR for local diffusion and bilinear discretization to create virtual points. These points and their features receive pillar features from texture and depth features via the outer product, and then through BEV pooling to generate the final image BEV features.}
  \label{fig:2}
  \vspace{-12pt}
\end{figure}

To address this, we propose a novel view transformation method that leverages the sparse depth information from LiDAR as a prior, diffuses within the same semantic class, followed by linear-increasing \cite{50} and linear-decreasing discretization to generate high-precision virtual point seeds around each co-visible point, as shown in \cref{fig:2}. First, we extract features from multi-camera images and generate semantic segmentation masks via a multi-task head, while simultaneously extracting image textual features and depth distribution weights, with the depth head and semantic head supplementing cross-task information through gated attention. To better utilize semantic information, we select 4x downsampled features for view transformation, as higher downsampling increases the semantic and depth ambiguity of pixels. 

Given the differences in sparsity between images and point clouds, we combine image semantic segmentation masks and sparse projected depth maps provided by LiDAR to diffuse depth values within the same semantic category masks, generating a semi-dense extended depth map. This process is as follows:
\begin{equation}
D_{\text{temp}}(i, j) = 
\frac{\sum\limits_{(p, q) \in N(i, j)} D(p, q) \cdot \mathbb{I}[M(p, q) = M(i, j)]}{\sum\limits_{(p, q) \in N(i, j)} \mathbb{I}[M(p, q) = M(i, j)]},
\end{equation}
where $N(i, j)$ represents the circle area with radius $r$ around the current point, and  \( M(i, j) \)  denotes the segmentation mask with N category. And $\mathbb{I}[M(p, q) = M(i, j)]$ checks if the semantic label at (p,q) matches that at (i,j), as follows:
\begin{equation}
\mathbb{I}[M(p, q) = M(i, j)] = 
\begin{cases} 
1, & \text{if } M(p, q) = M(i, j) \neq 0 \\
0, & \text{otherwise}
\end{cases}.
\end{equation}
And the final extended depth map \(D_(i, j) \) replaces the original co-points of the $D_{\text{temp}}(i, j)$.

Due to the projection deviation from 2D pixels to 3D points, we apply bidirectional linear incremental discretization to the extended depth map to obtain discrete virtual points, enhancing the accuracy of depth estimation. These steps reduce the number of virtual points, thereby improving inference speed. Finally, the image textual features $F_t$ and depth distribution weights $D_w$ are calculated by the outer product $F_t \otimes D_w$ to derive features for each virtual point and generate the BEV features ${F_{bev}^C}$ of the camera through BEV pooling. This method effectively integrates semantic information with sparse depth data, significantly enhancing the accuracy of pixel depth estimation and improving the speed of view transformations.
\subsection{Fusion-to-occupancy-driven Active Distillation}
\begin{figure}[t]
  \centering
  \includegraphics[width=1.0\linewidth,height=!]{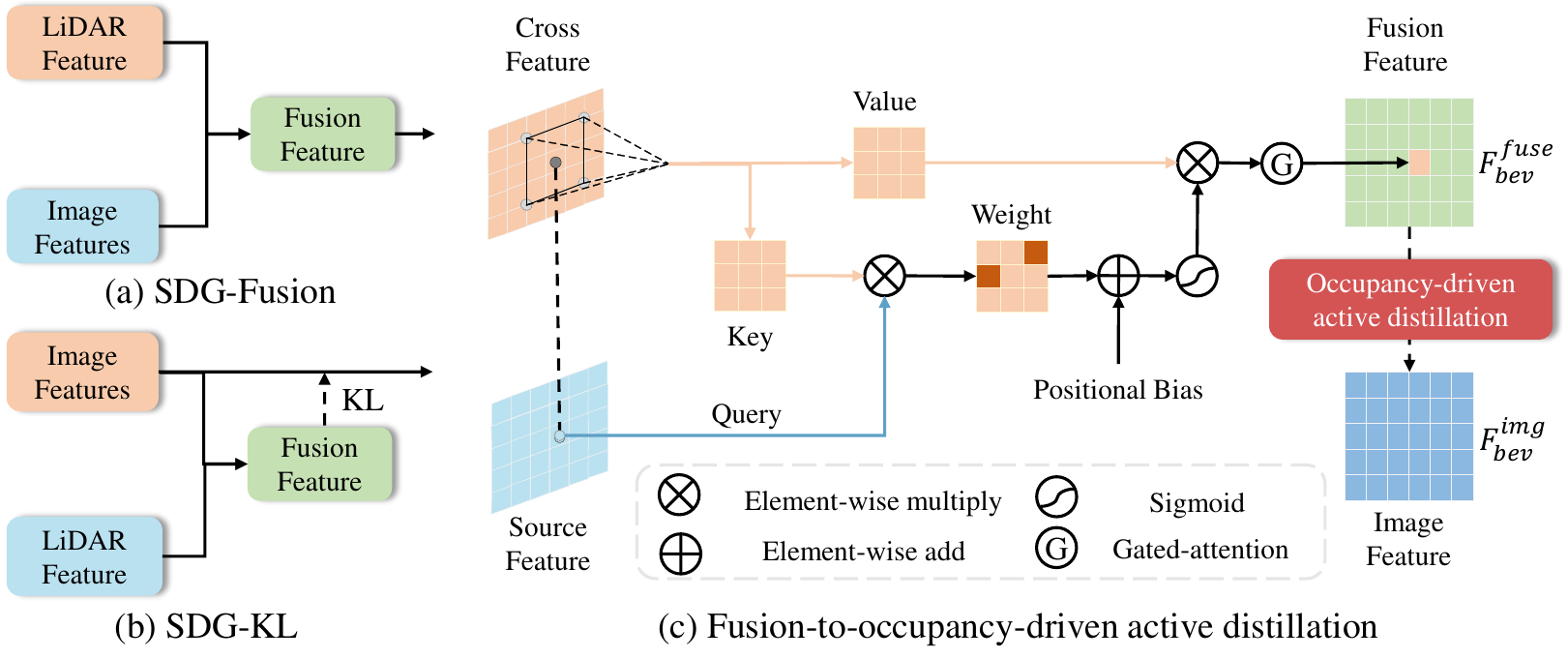}
  \caption{(a) The pipeline of SDG-Fusion. (b) The pipeline of SDG-KL. (c) The overall architecture of fusion-to-occupancy-driven active distillation. The source feature extracts neighborhood information from the corresponding pixels of the cross feature, where the source serves as the query, and the cross feature acts as both key and value for feature interaction. This interaction is dynamically refined through a gated attention mechanism to produce the fused feature. Subsequently, an occupancy-driven active distillation is used to unidirectionally integrate multimodal information into the image feature.}
  \label{fig:3}
  \vspace{-12pt}
\end{figure}
The LiDAR branch encompasses point cloud feature extraction, multimodal fusion, and occupancy-driven active distillation (as shown in \cref{fig:3}). Initially, the point cloud data undergoes voxelization and normalization to generate the initial features. We choose SPVCNN as our point-voxel feature encoder due to its efficiency in representing sparse point clouds while effectively preserving fine-grained details. Subsequently, we compress the voxel features at the corresponding scale to generate the BEV features ${F_{bev}^L}$.

The complementary information from LiDAR and cameras is critical for 3D perception. A naive fusion method typically concatenates LiDAR and image BEV features along the channel dimension to enhance performance. However, feature misalignment due to extrinsic conflicts \cite{41} limits the effectiveness of the fusion. Therefore, we propose a dynamic neighborhood feature fusion module. This module unidirectionally extracts neighborhood features from cross features and dynamically adjusts their weights into the source features using a gating attention mechanism. 

External projection deviations during the BEV feature construction process for LiDAR and images result in misalignment between LiDAR and camera BEV features \cite{41,42}. Therefore, we adopt neighborhood attention from \cite{43} to extract local patch features corresponding to the pixel from the cross features, and dynamically adjust the weights through gated attention to selectively enhance the fused feature representation. Specifically, the image features ${F_{bev}^C}$, as the source feature, are represented as a feature vector sequence $F_{img} \in \mathbb{R}^{{n \times m}}$, which is projected through a linear layer to obtain the query features $Q_{s}\in \mathbb{R}^{{n \times q}}$. Similarly, the LiDAR features ${F_{bev}^L}$ as cross feature are projected to obtain the key $K_{c} \in \mathbb{R}^{{n \times q}}$ and value $V_{c} \in \mathbb{R}^{{n \times v}}$features. The local neighborhood features \( F_{\text{neighbor}} \) for a query point \( i \) are computed by the following equation:
\begin{equation}
F_{\text{neighbor}} = \sigma \left( \frac{Q_{s}^i \cdot (K_c^{n(i)})^T + B(i, n(i))}{\sqrt{v}} \right) \cdot V_c^i ,
\end{equation}
where \( n(i) \) represent the neighborhood with the size of \( k \) centered at the same position in the cross feature, \( B(i, \rho(i)) \) denotes the relative positional biases and \( \sigma \) denotes to Softmax. For each pixel in the feature map, we calculate the local neighborhood features. The fused features ${F_{bev}^{fuse}}$ are then obtained from the local neighborhood features through gated attention, as follows:
\begin{equation}
F_{bev}^{fuse} = (\sigma (\text{Conv}( f_\text{Avg}(F_{\text{neighbor}}))) \cdot F_{\text{neighbor}},
\end{equation}
where $\sigma$ denotes the sigmoid function and Conv denotes linear transform matrix (e.g., 1x1 convolution), $f_\text{Avg}$ refers Adaptive Average Pooling. The fused features ${F_{bev}^{fuse}}$ are processed by the occupancy prediction head to obtain the SDG-Fusion model.

Additionally, to ensure real-time, we propose an occupancy-driven active distillation, where fused features are unidirectionally transferred to the image features. Specifically, LiDAR features are used as the source feature, while image features serve as the cross feature, resulting in LiDAR-dominant fused features. Inspired by \cite{51}, we then divide the space into two regions: the active region(AR), where both LiDAR and image features are occupied, and the inactive region(IR), where only LiDAR features are occupied. The details are as follows:
\begin{equation}
AR = \left( M_{\text{fused},i,j} = 1 \right) \land \left( M_{\text{img},i,j} = 1 \right),
\end{equation}
\begin{equation}
IR = \left( M_{\text{fused},i,j} = 1 \right) \land \left( M_{\text{img},i,j} = 0 \right),
\end{equation}
where a value of 1 in $M_{\text{mode},i,j}$ indicates that the coordinate is occupied by the respective modality.  Typically, the AR region is significantly larger than the IR region. To prevent the model from overemphasizing knowledge distillation in the AR region, we apply adaptive scaling based on the relative sizes of the AR and IR regions, as follows:
\begin{equation}
W_{I,i,j}^{(l_n)} =
\begin{cases}
\alpha, & \text{if } (i, j) \in AR, \\
\rho \times \beta, & \text{if } (i, j) \in IR, \\
0, & \text{otherwise},
\end{cases}
\end{equation}
where \( \rho = \frac{N_{AR}}{N_{IR}} \) represents the relative importance of the \( IR \) over \( AR \), \( \alpha \) and \( \beta \) are the intrinsic balancing parameters, and \( N_{AR} \) and \( N_{IR} \) are the number of pixels in \( AR \) and \( IR \), respectively. 

The distillation loss between BEV feature from teacher \( F^t \) and student \( F^s \) are:
\begin{equation}
L_{\text{distill}} = \sum_{c=1}^{C} \sum_{i=1}^{H} \sum_{j=1}^{W} W_{i,j} \left( F_{bev}^{fuse} - {F_{bev}^C} \right)^2.
\end{equation}

The network is trained with the sum of distillation and classification loss. The image features $F_{bev}^C$ are processed by the occupancy prediction head to obtain the SDG-KL model. 
\begin{table*}[t]
\footnotesize
    \begin{center}
    \resizebox{\textwidth}{!}{ 
    \setlength{\tabcolsep}{0.0057\linewidth}
	\begin{tabular}{l|c|c|c|c| c c c c c c c c c c c c c c c c c|c}
		\toprule
		Method
		& \rotatebox{90}{Input}& \rotatebox{90}{Backbone} & \rotatebox{90}{Visible Mask}  & mIoU
        & \rotatebox{90}{\textcolor{nothers}{$\blacksquare$} others}
		& \rotatebox{90}{\textcolor{nbarrier}{$\blacksquare$} barrier}
		& \rotatebox{90}{\textcolor{nbicycle}{$\blacksquare$} bicycle}
		& \rotatebox{90}{\textcolor{nbus}{$\blacksquare$} bus}
		& \rotatebox{90}{\textcolor{ncar}{$\blacksquare$} car}
		& \rotatebox{90}{\textcolor{nconstruct}{$\blacksquare$} const. veh.}
		& \rotatebox{90}{\textcolor{nmotor}{$\blacksquare$} motorcycle}
		& \rotatebox{90}{\textcolor{npedestrian}{$\blacksquare$} pedestrian}
		& \rotatebox{90}{\textcolor{ntraffic}{$\blacksquare$} traffic cone}
		& \rotatebox{90}{\textcolor{ntrailer}{$\blacksquare$} trailer}
		& \rotatebox{90}{\textcolor{ntruck}{$\blacksquare$} truck}
		& \rotatebox{90}{\textcolor{ndriveable}{$\blacksquare$} drive. suf.}
		& \rotatebox{90}{\textcolor{nother}{$\blacksquare$} other flat}
		& \rotatebox{90}{\textcolor{nsidewalk}{$\blacksquare$} sidewalk}
		& \rotatebox{90}{\textcolor{nterrain}{$\blacksquare$} terrain}
		& \rotatebox{90}{\textcolor{nmanmade}{$\blacksquare$} manmade}
		& \rotatebox{90}{\textcolor{nvegetation}{$\blacksquare$} vegetation}
            &\rotatebox{90}{Time(ms)}
		\\
		\midrule

        TPVFormer \cite{23}& {C}& R-50 & \ding{52} & 34.2 & 7.68 & 44.01 & 17.66 & 40.88 & 46.98 & 15.06 & 20.54 & 24.69 & 24.66 & 24.26 & 29.28 & 79.27 & 40.65 & 48.49 & 49.44 & 32.63 & 29.82&289.85 \\
        SurroundOcc \cite{24}& {C} & R-101 & \ding{52}  & 37.1 & 8.97 & 46.33 & 17.08 & 46.54 & 52.01 & 20.05 & 21.47 & 23.52 & 18.67 & 31.51 & 37.56 & 81.91 & 41.64 & 50.76 & 53.93 & 42.91 & 37.16&303.03 \\
        OccFormer \cite{44} & {C}& R-50 & \ding{52}  & 37.4 & 9.15 & 45.84 & 18.20 & 42.80 & 50.27 & 24.00 & 20.80 & 22.86 & 20.98 & 31.94 & 38.13 & 80.13 & 38.24 & 50.83 & 54.3 & 46.41 & 40.15&- \\
        VoxFormer \cite{25} & {C}& R-101 & \ding{52} & 40.7 & - & - & - & - & - & - & - & - & - & - & - & - & - & - & - & - & - &-\\
        FBOcc \cite{16}& {C}& R-50 & \ding{52}  & 42.1 & 14.30 & 49.71 & 30.0 & 46.62 & 51.54 & 29.3 & 29.13 & 29.35 & 30.48 & 34.97 & 39.36 & 83.07 & 47.16 & 55.62 & 59.88 & 44.89 & 39.58 &- \\ 
        PanoOcc \cite{45}& {C} & R-101 & - & 42.13 & 11.67 & 50.48 & 29.64 & 49.44 & 55.52 & 23.29 & 33.26 & 30.55 & 30.99 & 34.43 & 42.57 & 83.31 & 44.23 & 54.40 & 56.04 & 45.94 & 40.40&322.58 \\ 
        FastOcc \cite{31}& {C}& R-101 & \ding{52}  & 40.75 & 12.86 & 46.58 & 29.93 & 46.07 & 54.09 & 23.74 & 31.10 & 30.68 & 28.52 & 33.08 & 39.69 & 83.33 & 44.65 & 53.90 & 55.46 & 42.61 & 36.50&221.2 \\ 
        BEVDet4D \cite{46} & {C}& Swin-B & \ding{52}  & 42.5 & 12.37 & 50.15 & 26.97 & 51.86 & 54.65 & 28.38 & 28.96 & 29.02 & 28.28 & 37.05 & 42.52 & 82.55 & 43.15 & 54.87 & 58.33 & 48.78 & 43.79&1000.0 \\
        FlashOcc \cite{26}& {C}& Swin-B & \ding{52}  & 43.52 & 13.31 & 51.62 & 28.07 & 50.91 & 55.69 & 27.46 & 31.05 & 29.98 & 29.20 & 38.86 & 43.68 & 83.87 & 45.63 & 56.33 & 59.01 & 50.63 & 44.56&909.1 \\
        COTR \cite{32}& {C} & Swin-B & \ding{52}& 46.2 & \textbf{14.85} & 53.25 & \textbf{35.19} & 50.83 & 57.25 & 35.36 & 34.06 & 33.54 & \textbf{37.14} & 38.99 & 44.97 &84.46 & \textbf{48.73} & 57.60 & 61.08 & 51.61 & 46.72&840.34 \\
        HyDRa \cite{36}&{C+R}&R-50&-&44.40& - & - & - & 52.3 & 56.3 -& - & 35.9 & 35.10 & - & - & 44.1 &- &- & - & - & - & - & - \\
        OCCFusion \cite{47}&{C+L}&R-101&-&46.79&11.65&47.81&32.07&57.27&57.51&31.80&\textbf{40.11}&47.35&33.74&45.81&50.35&78.79&37.17&44.36&53.36&63.18&63.20&-\\
        RadOcc-LC \cite{48}& {C+L}& Swin-B & \ding{52}  & 49.38 & 10.93 & \textbf{58.23} & 25.01 & 57.89 & 62.85 & 34.04 & 33.45 & 50.07 & 32.05 & 48.87 & 52.11 & 82.90 & 42.73 & 55.27 & 58.34 & 68.64 & 66.01&3333 \\
	\bottomrule
         SDG-KL & {C+L}& R-50 & \ding{52}  &50.16 & 12.26 & 57.12 & 23.69 & 58.77 & 62.74 & 34.55 & 36.19 & 50.1 & 32.05 & 49.89 & 51.24  & 84.1 & 46.05 & 57.2 & 61.45 & 69.56&65.78 &\textbf{83}\\
        SDG-Fusion & {C+L}& R-50 & \ding{52} &\textbf{51.66} & 13.21 & 57.77 & 24.3 & \textbf{60.33} & \textbf{64.28} & \textbf{36.21} & 39.44 & \textbf{52.36} & 35.80 & \textbf{50.91} & \textbf{53.65} &\textbf{84.56} & 47.45 & \textbf{58.00} & \textbf{61.61} & \textbf{70.67} & \textbf{67.65} &133\\
        \bottomrule
	\end{tabular}}
 \vspace{-8pt}
 \caption{3D Occupancy prediction performance on the Occ3D-nuScenes dataset. We present the  mean IoU over categories and the IoUs for different classes. The best scores for each class are highlighted in bold. In the Input, the C, L, and R denote camera, LiDAR, and radar, respectively. In the backbone, R represents ResNet, while Swin stands for Swin Transformer.}
 \label{tab:sota}
    \end{center}
    \vspace{-12pt}
\end{table*}
\begin{table*}[ht]
\footnotesize
    \begin{center}
    \resizebox{\textwidth}{!}{ 
    \setlength{\tabcolsep}{0.0057\linewidth}
	\begin{tabular}{l|c|c|c| c c c c c c c c c c c c c c c c c}
		\toprule
		Method
		& \rotatebox{90}{Input}& \rotatebox{90}{Backbone}   & mIoU
		& \rotatebox{90}{\textcolor{nbarrier}{$\blacksquare$} barrier}
		& \rotatebox{90}{\textcolor{nbicycle}{$\blacksquare$} bicycle}
		& \rotatebox{90}{\textcolor{nbus}{$\blacksquare$} bus}
		& \rotatebox{90}{\textcolor{ncar}{$\blacksquare$} car}
		& \rotatebox{90}{\textcolor{nconstruct}{$\blacksquare$} const. veh.}
		& \rotatebox{90}{\textcolor{nmotor}{$\blacksquare$} motorcycle}
		& \rotatebox{90}{\textcolor{npedestrian}{$\blacksquare$} pedestrian}
		& \rotatebox{90}{\textcolor{ntraffic}{$\blacksquare$} traffic cone}
		& \rotatebox{90}{\textcolor{ntrailer}{$\blacksquare$} trailer}
		& \rotatebox{90}{\textcolor{ntruck}{$\blacksquare$} truck}
		& \rotatebox{90}{\textcolor{ndriveable}{$\blacksquare$} drive. suf.}
		& \rotatebox{90}{\textcolor{nother}{$\blacksquare$} other flat}
		& \rotatebox{90}{\textcolor{nsidewalk}{$\blacksquare$} sidewalk}
		& \rotatebox{90}{\textcolor{nterrain}{$\blacksquare$} terrain}
		& \rotatebox{90}{\textcolor{nmanmade}{$\blacksquare$} manmade}
		& \rotatebox{90}{\textcolor{nvegetation}{$\blacksquare$} vegetation}
		\\
		\midrule
    SurroundOcc \cite{24}& C& R-101   & 20.3    & 20.5 & 11.6 & 28.1 & 30.8 & 10.7 & 15.1 & 14.0 & 12.0 & 14.3 & 22.2 & 37.2 & 23.7 & 24.4 & 22.7 & 14.8 & 21.8 \\ 
    OccFormer \cite{44} & C & R-101 & 20.1  & 21.1 & 11.3 & 28.2 & 30.3 & 10.6 & 15.7 & 14.4 & 11.2 & 14.0 & 22.6 & 37.3 & 22.4 & 24.9 & 23.5 & 15.2 & 21.1 \\
    C-CONet \cite{14}& C & R-101 & 18.4    & 18.6 & 10.0 & 26.4 & 27.4 & 8.6 & 15.7 & 13.3 & 9.7 & 10.9 & 20.2 & 33.0 & 20.7 & 21.4 & 21.8 & 14.7 & 21.3 \\
    FB-Occ \cite{16}& C& R-101   & 19.6   & 20.6 & 11.3 & 26.9 & 29.8 & 10.4 & 13.6 & 13.7 & 11.4 & 11.5 & 20.6 & 38.2 & 21.5 & 24.6 & 22.7 & 14.8 & 21.6 \\
    RenderOcc \cite{48} & C & R-101   & 19.0    & 19.7 & 11.2 & 28.1 & 28.2 & 9.8 & 14.7 & 11.8 & 11.9 & 13.1 & 20.1 & 33.2 & 21.3 & 22.6 & 22.3 & 15.3 & 20.9 \\
    L-CONet  \cite{14} & L & -  & 17.7 & 19.2 & 4.0 & 15.1 & 26.9 & 6.2 & 3.8 & 6.8 & 6.0 & 14.1 & 13.1 & 39.7 & 19.1 & 24.0 & 23.9 & 25.1 & 35.7 \\
        FlashOcc* \cite{26} & C  & R-50 & 44.1 &44.2 & 11.0 & 54.1 & 60.5 & 26.1 & 22.6 & 31.3 & 15.3 & 38.8 & 47.1 & 80.5 & 42.0 &48.2 & 53.7 & 60.8 & 70.0 \\
    M-CONet \cite{14}& C+L & R-101  & 24.7  & 24.8 & 13.0 & 31.6 & 34.8 & 14.6 & 18.0 & 20.0 & 14.7 & 20.0 & 26.6 & 39.2 & 22.8 & 26.1 & 26.0 & 26.0 & 37.1 \\
    Co-Occ \cite{34} & C+L & R-101   & 27.1   & 28.1 & 16.1 & 34.0 & 37.2 & 17.0 & 21.6 & 20.8 & 15.9 & 21.9 & 28.7 & 42.3 & 25.4 & 29.1 & 28.6 & 28.2 & 38.0 \\
    OccFusion \cite{47}& C+L+R & R-101 & 27.3    & 27.1 & 19.6& 33.7 & 36.2 & 21.7 & 24.8 & 25.3 & 16.3 & 21.8 & 30.0 & 39.5 & 19.9 & 24.9 & 26.5 & 28.9 & 40.4 \\
    DAOcc \cite{49} & C+L  & R-50 & 30.5 &30.8 & 19.5 & 34.0 & 38.8 & 25.0 &27.7 & 29.9 & 22.5 & 23.2 & 31.6 & 41.0 & 25.9 & 29.4 & 29.9 & 35.2 & 43.5 \\
    \bottomrule
        SDG-Fusion & {C+L}& R-50& 31.7 &31.2 & 15.6 & 39.7 & 42.4 & 22.2 &26.9& 29.7 & 22.7 & 24.2 & 33.2 & 45.9 & 25.1& 31.8 & 33.5 & 38.9 & 44.5 \\
    SDG-KL* & {C+L}& R-50  &50.4 & 51.4  &\textbf{20.9} & 59.6 & 65.8 & \textbf{34.1} & 33.1 & 49.0 &\textbf{23.4} & 45.1 &53.7& 76.5& 44.0 & 53.5 & 58.2 & 65.2&72.9 \\
    SDG-Fusion*& {C+L}& R-50  &\textbf{52.2}  & \textbf{54.3} & 16.0 & \textbf{61.4} & \textbf{66.9} & 33.2 & \textbf{33.5}& \textbf{49.2} & 22.2 & \textbf{47.8} & \textbf{54.6} &\textbf{83.9} & \textbf{49.4}& \textbf{55.8} & \textbf{60.2} & \textbf{70.2} & \textbf{76.5} \\
    \bottomrule
	\end{tabular}}
 \vspace{-8pt}

 \caption{3D Occupancy prediction performance on the SurroundOcc-nuScenes validation set. The best scores for each class are highlighted in bold. In the Input, the C, L, and R denote camera, LiDAR, and radar, respectively. * means the performance is achieved through a visible mask similar to \cite{13}.}
     \label{tab:surroundocc}
    \end{center}
    \vspace{-16pt}
\end{table*}
\subsection{Occupancy Prediction}
To obtain 3D prediction outputs from coarse BEV features generated by view transformation, we propose an occupancy prediction system consisting of a BEV feature encoder and an occupancy prediction head. The BEV encoder uses several residual blocks for multi-scale feature diffusion and integrates a feature pyramid to acquire BEV features at the target scale. The occupancy prediction head extracts global features with multiple 3x3 convolutional layers and includes a channel-to-height transformation module. This module reshapes the BEV features from a ${F}_{out} \in \mathbb{R}^{B \times C \times H \times W}$ to ${F}_{final} \in \mathbb{R}^{B \times {C}_{N} \times D \times H \times W}$, where $B, C, W, H$, and $D$ represent the batch size, channel number, class number, and the dimensions of the 3D space, respectively, with $C$ = ${C}_{N} \times D$. Compared to traditional 3D encoders and occupancy prediction heads, this design significantly improves speed while maintaining comparable performance.

%% file: sec/4_Experiments.tex
\section{Experiments}
We conduct experiments on the large-scale benchmark dataset Occ3D-nuScenes to validate the efficacy of our proposed methods. Additionally, we conduct ablation experiments to verify the effectiveness of each component in our method.
\subsection{Datasets}
Occ3D-nuScenes  \cite{13} is a large-scale autonomous driving dataset, which includes 1,000 urban traffic scenes under various conditions, the data is split into 700 training, 150 validation, and 150 testing scenes. The occupancy grid is defined within a range of -40m to 40m along the X and Y axes and -1m to 5.4m along the Z axis. The voxel size for occupancy labeling is set to 0.4m × 0.4m × 0.4m. The semantic labels include 17 categories consisting of 16 known object classes with an additional 'empty' class. Compared to Occ3D-nuScenes, SurroundOcc \cite{24} is also based on the nuScenes dataset but its prediction range is from -50m to 50m for X and Y axes, and -5m to 3m along the Z axis, with the voxel label size of 0.5m × 0.5m × 0.5m.
 \begin{figure*}[t!]
\centering
\includegraphics[width=0.92\textwidth,height=!]{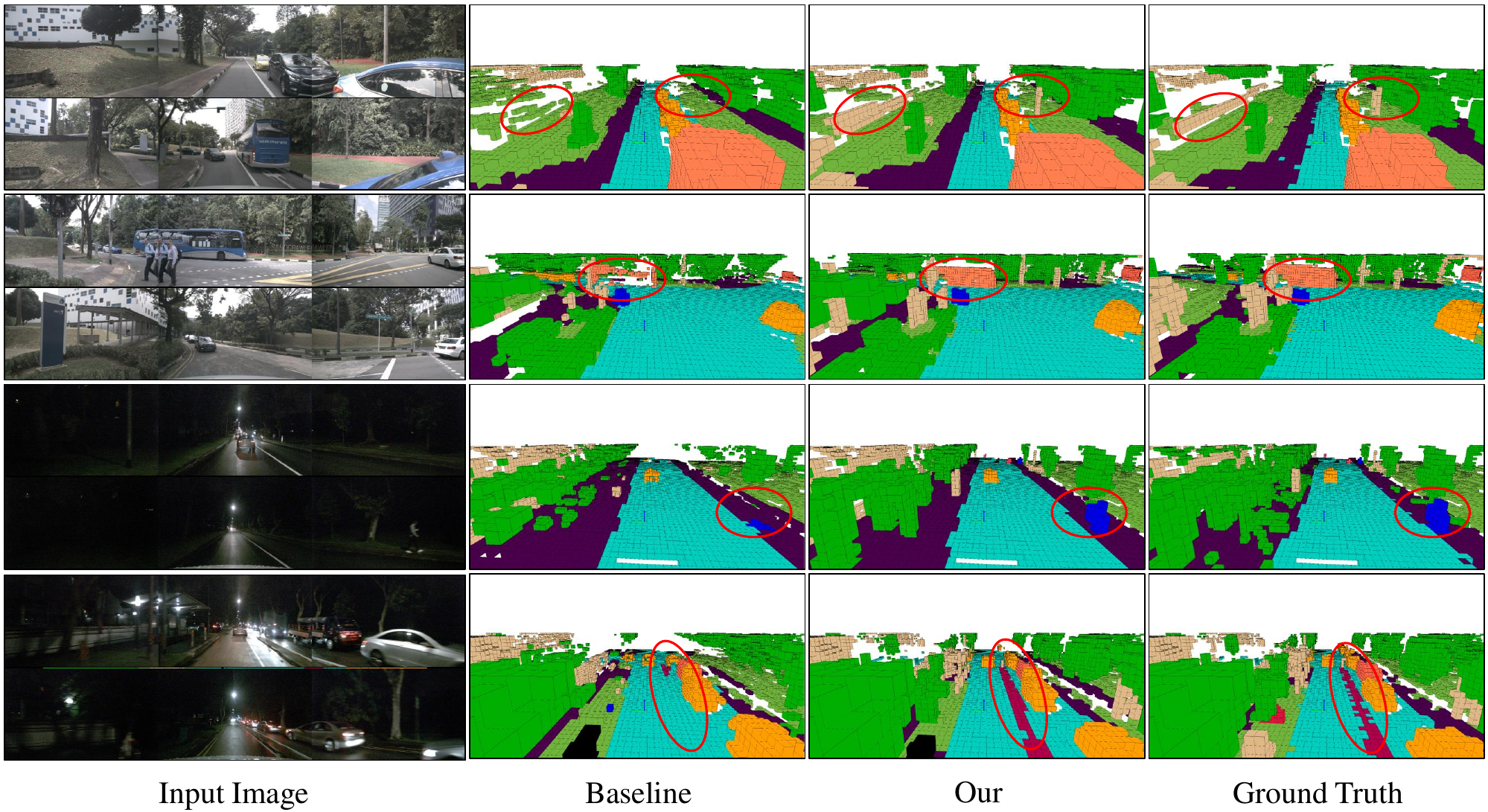}
\vspace{-8pt}
\caption{Qualitative results of SDG on the validation set of Occ3D-nuScenes. Each pair of rows displays results from day and low-light scene, respectively. Within each row, images from left to right represent the input images, baseline, our results, and the ground truth.}
\label{keshihua}
\vspace{-12pt}
\end{figure*}
\subsection{Implementation Details}
We use ResNet-50 as the default image backbone and SPVCNN as the LiDAR backbone. The model is trained on a GeForce RTX 4090 GPU using the AdamW optimizer with a learning rate of 1e-4 and gradient clipping. For semantic and depth-guided visual transformations, the bilinear incremental discretization range and the number of diffusion feature layers are set to 1 m and 8, respectively.
\subsection{Comparing with SOTA methods}
\begin{figure}[t]
\centering\includegraphics[width=0.41\textwidth]{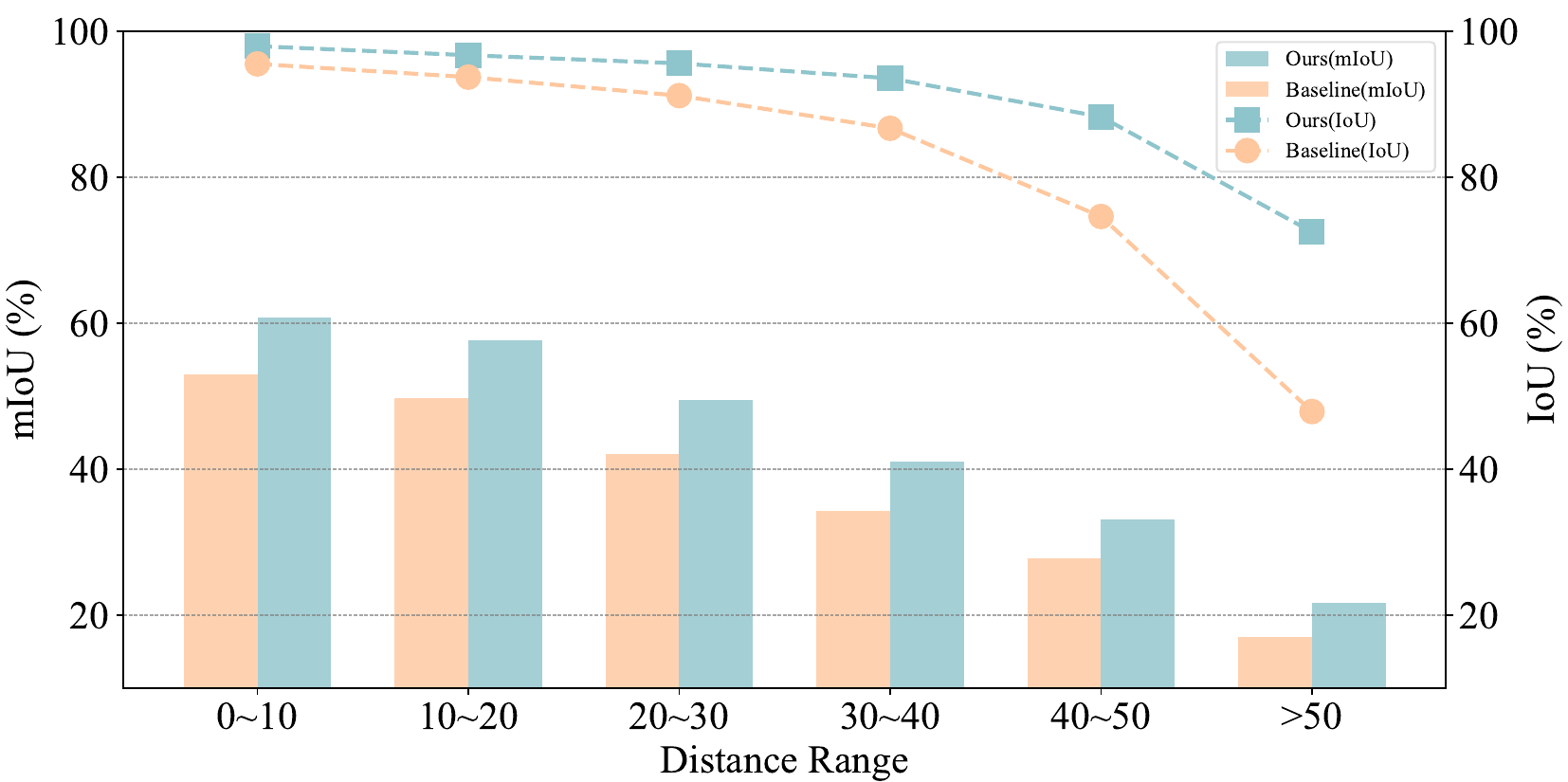}
\vspace{-8pt}
\caption{ Distance-based evaluation on Occ3D-nuScenes. As the distance increases, the point cloud becomes sparse.}
\label{dis}
\vspace{-12pt}
\end{figure}
\textbf{Occ3D-nuScenes.} As shown in Table \ref{tab:sota}, we report the quantitative comparison of existing state-of-the-art methods for 3D occupancy prediction tasks on Occ3D-nuScenes. Most existing approaches are predominantly based on camera-only algorithms, with relatively few focusing on multi-sensor fusion. Our method, which employs a compact backbone and a lightweight LiDAR branch, achieves state-of-the-art performance in terms of mIoU and the majority of class-wise IoUs. Additionally, our approach achieves the best inference speed, meeting the real-time requirements of autonomous driving scenarios. The visualization on Occ3D-nuScenes validation set is shown in \cref{keshihua}. Compared to the baseline, our method effectively identifies categories that the baseline fails to correctly predict in both day and night scenarios. More visualizations  can be found in supplementary material.

\textbf{SurroundOcc-nuScenes.} Table \ref{tab:surroundocc} provides a quantitative comparison on the SurroundOcc validation set, highlighting the performance of our method, SDG-OCC, compared to other approaches. Using both LiDAR and camera inputs, our method achieves SOTA performance on the SurroundOcc validation set, even over larger distances. This success is attributed to our semantic and depth-guided view transformation, which enhances depth estimation accuracy and enables robust occupancy prediction across varying grid sizes and distances. Furthermore, we utilize visual masks generated from \cite{14}, achieving performance comparable to OCC3d-nuscenes. Notably, our method uses only a lightweight ResNet50 backbone and a lower resolution of 256×704, underscoring its effectiveness and efficiency.

\textbf{Analysis of results within different ranges.} We further evaluate different ranges surrounding the car to provide a comprehensive analysis. \cref{dis} clearly illustrates our mIoU and iou relative to the baseline FlashOcc. Short-range understanding is critical due to the limited reaction time for autonomous vehicles. Our method significantly outperforms the baseline in both mIoU and IoU. In long-range areas, where LiDAR data is sparse and few pixels define the depth of large regions, our approach still achieves superior IoU performance.
\begin{table}[t]
\centering
\resizebox{0.85\linewidth}{!}{
\begin{tabular}{c ccc c}
\hline
Baseline & SDG & FOAD & iou(\%) & mIoU(\%) \\  
\hline
\checkmark &          &          & 90.27 & 37.84  \\
\checkmark & \checkmark &          & 94.62 & 48.51  \\
\checkmark &          & \checkmark & 94.76 & 44.92  \\
\checkmark & \checkmark & \checkmark & 95.35 & 51.66 \\
\hline
\end{tabular}}
\vspace{-4pt}
\caption{Ablation study on Occ3D-nuScenes dataset with SDG-Fusion. SDG: Semantic and Depth-Guided View Transformations. FOAD: Fusion-To-Occupancy-Driven Active Distillation.}
\label{AS}
\vspace{-16pt}
\end{table}
\begin{table}[t]
\centering
\setlength{\tabcolsep}{16pt} 
\begin{tabular}{cc|c|c}
\hline
$r$ & $l$ & IoU(\%) & mIoU (\%) \\
\hline
1 & 4 & 95.34  & 51.15 \\
1 & 8 & 95.35 & \textbf{51.66} \\
1 & 12 & 95.38  & 51.28 \\
2 & 4 & 95.35 & 50.4 \\
2 & 8 & \textbf{95.39} & 51.03 \\
2 & 12 & 95.30  & 51.12 \\
\hline
\end{tabular}
\vspace{-6pt}
\caption{Ablation study of the hyperparameter used in SDG view transformation module on Occ3D-nuScenes.}
\label{VT}
\vspace{-20pt}
\end{table}
\begin{table}[t]
\centering
\setlength{\tabcolsep}{10pt} 
\begin{tabular}{c|cccc}
\hline 
\textbf{$k$}  & \textbf{3} & \textbf{5} & \textbf{7} & \textbf{9} \\
\hline
IoU (\%)  & 95.32 & \textbf{95.40} & 95.35 & 95.26 \\
\hline
mIoU (\%)  & 51.22 & 51.40& \textbf{51.66} & 51.08 \\
\hline
\end{tabular}
\vspace{-4pt}
\caption{Ablation study of the hyperparameter used in feature fusion of FOAD on Occ3D-nuScenes. The K denotes the size of the pixel region corresponding to the neighborhood feature extraction.}
\label{tab:4}
\vspace{-17pt}
\end{table}
\subsection{Ablation study}
\textbf{The Effectiveness of Each Component.} The results are shown in Table \ref{AS}, we can observe that all components make their own performance contributions. The baseline achieves 90.27\% of IoU and 37.84\% of mIoU. We first integrated the Semantic and Depth-Guided (SDG) View Transformation into the baseline model, which brings 4.35\% and 10.67\% performance gain in IoU and mIoU. Fusion enhanced by integrating additional LiDAR information, the IoU and mIoU have been significantly improved by 4.49\% and 7.08\%, respectively. By using both SDG and Fusion, outperforming the baseline by 5.19\% of IoU and 13.82\% of mIou. 

\textbf{The Effectiveness of SDG View Transformation.} To further demonstrate the effect of SDG View Transformation, we conducted hyperparameter analysis experiments. In the SDG view transformation, the range $r$ of bilinear growth discretization and the diffusion feature layers $l$ control the virtual point generation of SDG. As shown in Table \ref{VT}, lower depth precision (e.g., $r=2$ and $l=4$) results in slightly reduced performance compared to other configurations. However, excessive depth precision does not lead to additional gains, with the optimal performance observed at $r=1$ and $l=8$.

\textbf{The Effectiveness of FOAD Module.} We perform a hyperparameter analysis of the FOAD module. For neighborhood feature fusion, the parameter $K$ controls the fusion of features from neighboring pixels. As shown in Table \ref{tab:4}, increasing $K$ does not consistently improve performance, with optimal results achieved at $K=7$.

%% file: sec/5_conclusion.tex
\section{Conclusion}
\label{sec:conclusion}
In this paper, we introduce a multimodal 3D semantic occupancy prediction framework, termed SDG-OCC, designed to achieve higher accuracy and competitive inference speed by fusing LiDAR information in the BEV perspective. To address the inaccurate depth estimation in view transformations, we propose a semantic and depth-guided view transformation method. This approach integrates pixel semantics and corresponding point depth through diffusion and bilinear discretization, effectively reducing invalid image features and significantly enhancing the speed and accuracy of semantic occupancy. Meanwhile, We propose a fusion-to-occupancy-driven active distillation that incorporates multimodal features and selectively transfers multimodal knowledge to image features based on LiDAR-identified regions. Our method achieves the SOTA performance with real-time processing on the Occ3D-nuScenes dataset and comparable performance on the more challenging SurroundOcc-nuScenes dataset, demonstrating its effectiveness.
